\documentclass[a4paper,conference]{IEEEtran}
\ifCLASSINFOpdf
  \usepackage[pdftex]{graphicx}
\else
\fi
\usepackage{multirow}
\usepackage[normalem]{ulem}
\useunder{\uline}{\ul}{}

\usepackage{amsmath}
\begin{document}
%
\title{Accurate 3-D Reconstruction with RGB-D Cameras using Depth Map Fusion and Pose Refinement}

\author{\IEEEauthorblockN{Markus Ylim\"aki and Janne Heikkil\"a}
\IEEEauthorblockA{Center for Computer Vision and Signal Analysis\\
University of Oulu\\
Oulu, Finland\\
Email: firstname.lastname@oulu.fi}
\and
\IEEEauthorblockN{Juho Kannala}
\IEEEauthorblockA{Department of Computer Science\\
Aalto University\\
Espoo, Finland\\
Email: juho.kannala@aalto.fi}}


%


\maketitle

\begin{abstract}
Depth map fusion is an essential part in both stereo and RGB-D based 3-D reconstruction pipelines. Whether produced with a passive stereo reconstruction or using an active depth sensor, such as Microsoft Kinect, the depth maps have noise and may have poor initial registration. In this paper, we introduce a method which is capable of handling outliers, and especially, even significant registration errors. The proposed method first fuses a sequence of depth maps into a single non-redundant point cloud so that the redundant points are merged together by giving more weight to more certain measurements. Then, the original depth maps are re-registered to the fused point cloud to refine the original camera extrinsic parameters. The fusion is then performed again with the refined extrinsic parameters. This procedure is repeated until the result is satisfying or no significant changes happen between iterations. The method is robust to outliers and erroneous depth measurements as well as even significant depth map registration errors due to inaccurate initial camera poses.
\end{abstract}


%
\IEEEpeerreviewmaketitle

\section{Introduction}

The three-dimensional (3-D) reconstruction of a scene or an object is a classical problem in computer vision \cite{Seitz06}. The reconstruction methods can be roughly categorized into passive stereo approaches (e.g. \cite{Furukawa10a,Fuhrmann14,Ylimaki15a}) and RGB-D based methods using an active depth sensor (e.g. \cite{Newcombe11,Whelan12,Niessner13}). The stereo approaches reconstruct the scene purely from photographs while the RGB-D methods use a specific depth sensor, such as Microsoft Kinect, to provide depth measurements of the scene.

In the depth map based stereo reconstruction methods, such as \cite{Merrell07,Li10,Tola12,Fuhrmann14}, and especially in the RGB-D reconstruction, the fusion of depth maps is an essential part of the modeling pipeline and may have a significant influence on the final result. The simplest way to fuse depth maps is to register them into the same coordinate system but this approach will lead to a huge number of redundant points, which makes the further processing very slow.

A better way is to aim directly at a non-redundant point cloud so that overlapping points from different depth maps do not increase the redundancy of the point cloud \cite{Kyostila13,Ylimaki17}. In this paper, we propose a method which is able to reconstruct non-redundant point clouds from redundant, noisy and poorly registered depth maps. That is, at first, the method merges a sequence of depth maps into a single non-redundant point cloud so that new measurements are either added to the cloud or used to refine the nearby existing points. Then, the method re-registers the original depth maps into the fused point cloud to refine the camera poses and repeats the fusion step. The experiments show that the proposed method significantly reduces the amount of outliers and, especially, badly registered points in the final point cloud.

\begin{figure}[t!]
\centering
\includegraphics[width=0.48\columnwidth]{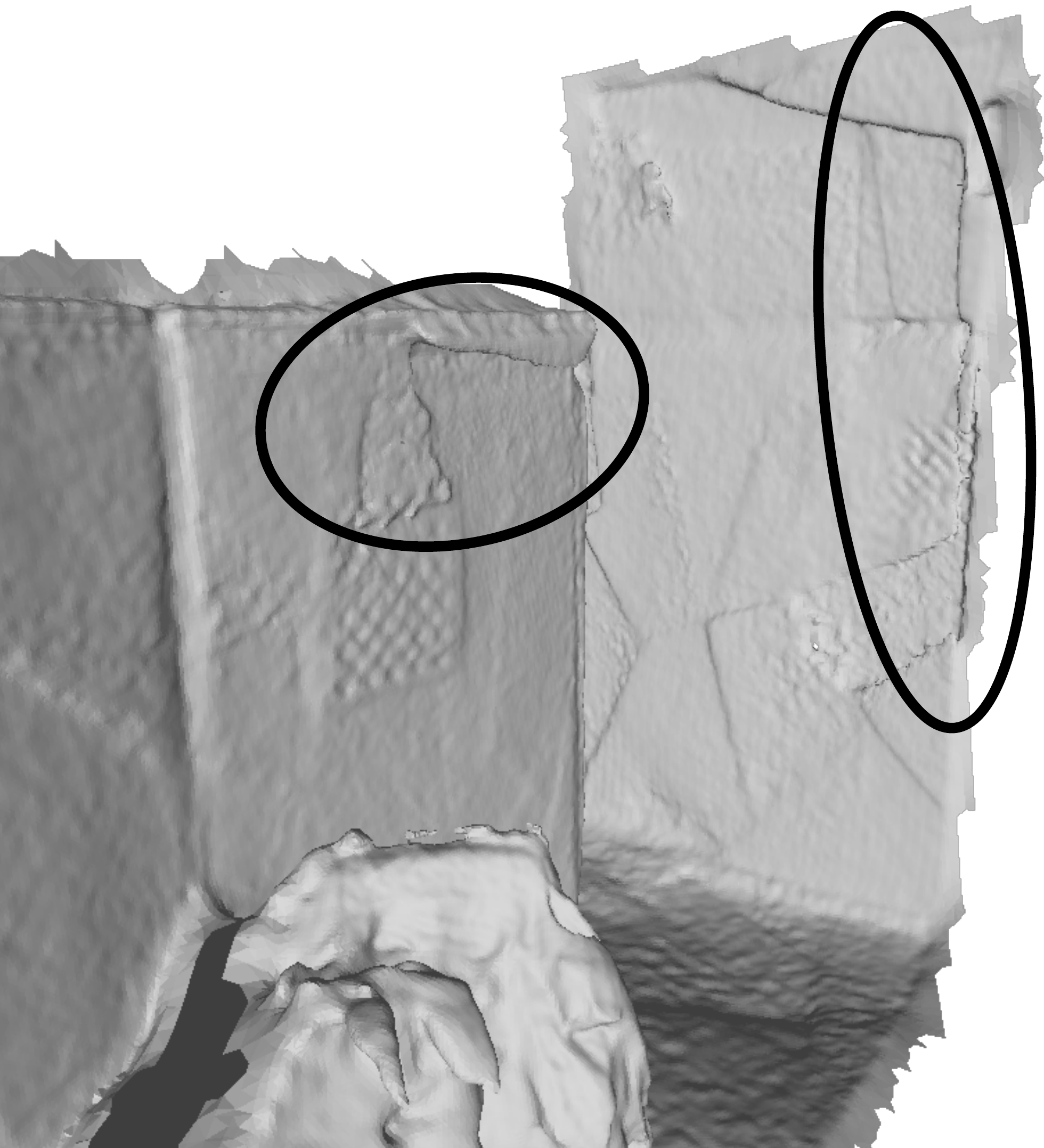}
\includegraphics[width=0.48\columnwidth]{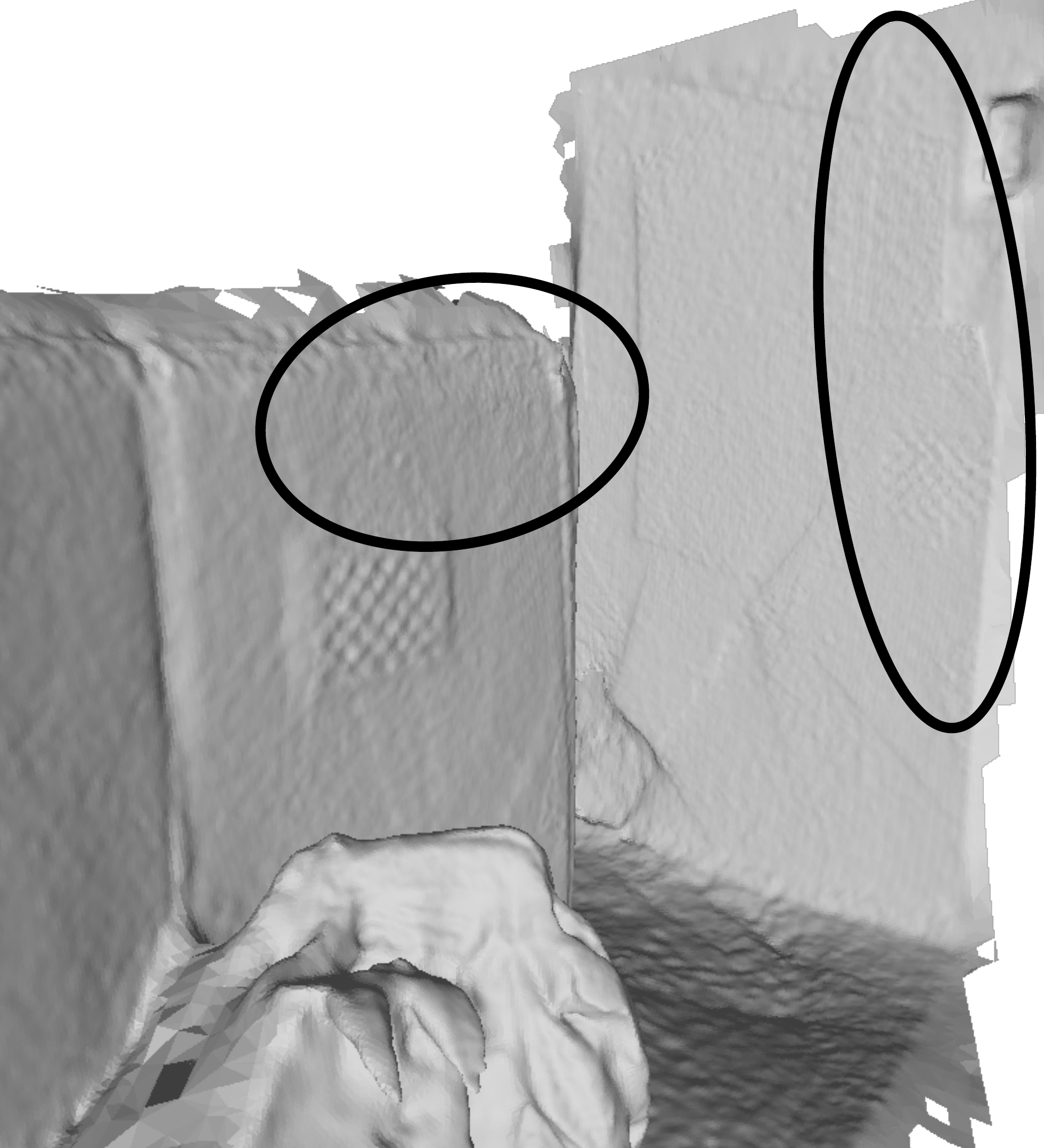}
\caption{Parts of the mesh reconstructions of Office2 made with Poisson Surface Reconstruction \cite{Kazhdan06} from the point clouds made with the method in \cite{Ylimaki17} (left) and the proposed one (right). The proposed method reduces the ambiguity of the surfaces, and thus, produces more accurate results (black ellipses).}
\label{mesh_overview}
\end{figure}

Figure \ref{mesh_overview} presents parts of the triangle meshes created using Poisson Surface Reconstruction (PSR) \cite{Kazhdan06} from the point cloud created with the method in \cite{Ylimaki17} and with the proposed method. The figure clearly shows the difference between the results. Inaccurate registration of the depth maps cause ambiguous surfaces in the point cloud as well as in the mesh. The proposed method is able to produce results where the reconstructed surfaces are smoother and less ambiguous (black ellipses).

The rest of the paper is organized as follows. Section \ref{related_work} introduces some previous works and their relationship to our approach which is described in more detail in Section \ref{method}. The experimental results are presented and discussed in Section \ref{results} and Section \ref{conclusion} concludes the paper.

\section{Related work} \label{related_work}

Fusion of depth maps as a part of a 3-D reconstruction has been widely studied both in passive stereo reconstruction pipelines \cite{Goesele06,Merrell07,Zach07,Li10,Tola12} as well as in RGB-D based methods \cite{Newcombe11,Whelan12,Roth12,Niessner13,Choi15}. Some of the existing methods are shortly described below.

Goesele et al. \cite{Goesele06} merge the depth maps using the volumetric approach in \cite{Curless96}, which uses a directional uncertainty for the depth measurements captured with a range scanner. However, the uncertainty in \cite{Goesele06} is based on the photometric consistency of the depth measurements among the views where they are visible, and therefore, the uncertainty is not directly depth dependent. Later, their work has led to publicly available reconstruction application called Multi-View Environment \cite{Fuhrmann14} where the depth map fusion is still based on \cite{Curless96} but enhanced to work with depth maps having multiple scales.

The depth map fusion presented in \cite{Merrell07} is able to reconstruct a scene from a live video in real time. Their method uses simple visibility violation rules but also exploits confidence measures for the points. However, the confidence is based on image gradients, and thus, is not depth dependent either.

Zach et al. have proposed a depth map fusion which incorporates a minimization of an energy function consisting of a total variation (TV) as a regularization term and a $L^1$ norm as a data fidelity term \cite{Zach07}. The terms are based on signed distance fields computed similarly as in \cite{Curless96}. 

Li et al. \cite{Li10} have utilized the bundle adjustment approach in the depth map fusion. Again, their method relies strongly on the photo consistency of matched pixels between stereo images, and thus, the method may not work that well in RGB-D based reconstruction approaches like \cite{Kyostila13} and \cite{Ylimaki17}. 

Being relatively simple, the fusion in \cite{Tola12} can be applied both in passive stereo and RGB-D based approaches. Similarly to \cite{Kyostila13} and \cite{Ylimaki17}, their method produces a non-redundant point cloud out of a set of overlapping depth maps. Nevertheless, they simply preserve points which do not have redundancy or have the highest accuracy among redundant points and do not exploit any uncertainty based merging of the redundant points.

As a summary, the passive stereo reconstruction approaches, described above, do not exploit any depth dependent uncertainty for the depth measurements like in \cite{Kyostila13} and \cite{Ylimaki17} and the confidence measures are naturally very often based on the photometric consistence. In addition, they usually assume noisy depth maps, similarly to \cite{Ylimaki17}, but relatively accurate camera poses.

The interest towards RGB-D based approaches has been increasing widely since Microsoft released the first generation Kinect device (Kinect V1) in 2010. The RGB-D reconstruction algorithms, such as KinectFusion presented in \cite{Newcombe11}, are known as real-time approaches producing scale and resolution limited reconstructions because of the memory consuming voxel based representation of the models. 

KinectFusion is able to reconstruct a scene inside a fixed volume in real-time. The registration of the depth maps as well as the camera pose estimation is based on the iterative closest point (ICP) algorithm. Whelan et al. \cite{Whelan12} and Roth \& Vona \cite{Roth12} have provided extensions to KinectFusion which allow larger reconstructions but are still quite scale and resolution limited.

The memory restriction has been avoided especially by Nie{\ss}ner et al. \cite{Niessner13} who proposed a method where the surface data can be efficiently streamed in or out of a specific hash table. However, as for all methods designed for live video reconstruction, their method may not work that well with depth maps having wide baselines.


Point cloud based RGB-D reconstruction approaches do not have similar scale or resolution limitation than the voxel based approaches. One of such methods was proposed by Ky\"ostil\"a et al. in \cite{Kyostila13} where the point cloud is obtained by merging a sequence of depth maps iteratively. That is, the method loops through every pixel in every registered depth map in the sequence and either adds the point into the cloud of points in the space, if there are no other points nearby, or uses the measurement to refine an existing point. This way, the fusion does not increase the redundancy of the cloud and the location uncertainty of each point guarantees that the refinement takes all redundant measurements into account \cite{Kyostila13} and not just preserve the one which seems to be the most accurate \cite{Tola12}. The uncertainty is based on empirically defined, depth depended variances.

Ky\"ostil\"a's method is designed for Kinect V1 and the main contribution of their work is the fusion of redundant depth maps. Thus, their method is not very robust to outliers or certain measurement or registration errors. Ylim\"aki et al. recently improved the method in order to make it work with the newer Kinect device (Kinect V2) and provided three extensions to boost its robustness and accuracy \cite{Ylimaki17}. The extensions include depth map pre-filtering to reduce the amount of outliers, improved uncertainty covariance to compensate for the measurement variances and make the method more accurate and filtering of the final point cloud to reduce the amount of erroneous measurements. 


Although the method in \cite{Ylimaki17} is able to reduce the amount of outliers quite significantly, it cannot handle severe registration errors which may cause ambiguous and rough surfaces both in the point cloud but also in the meshed model build with PSR \cite{Kazhdan06}, for instance (see Fig. \ref{mesh_overview}).

In this paper, to overcome the above limitation of \cite{Ylimaki17}, we further develop the method with a re-registration extension. That is, after acquiring the first fused point cloud, our method first registers the original depth maps with the point cloud and then, like in KinectFusion \cite{Newcombe11}, refines the camera poses according to the new registration, and finally, reruns the fusion. In the experimental results as well as in Figure \ref{mesh_overview}, we show that the proposed method produces significantly better results robustly, than presented in \cite{Kyostila13} and \cite{Ylimaki17}.


\section{Method} \label{method}

\begin{figure}[t!]
\centering
\includegraphics[width=0.95\columnwidth]{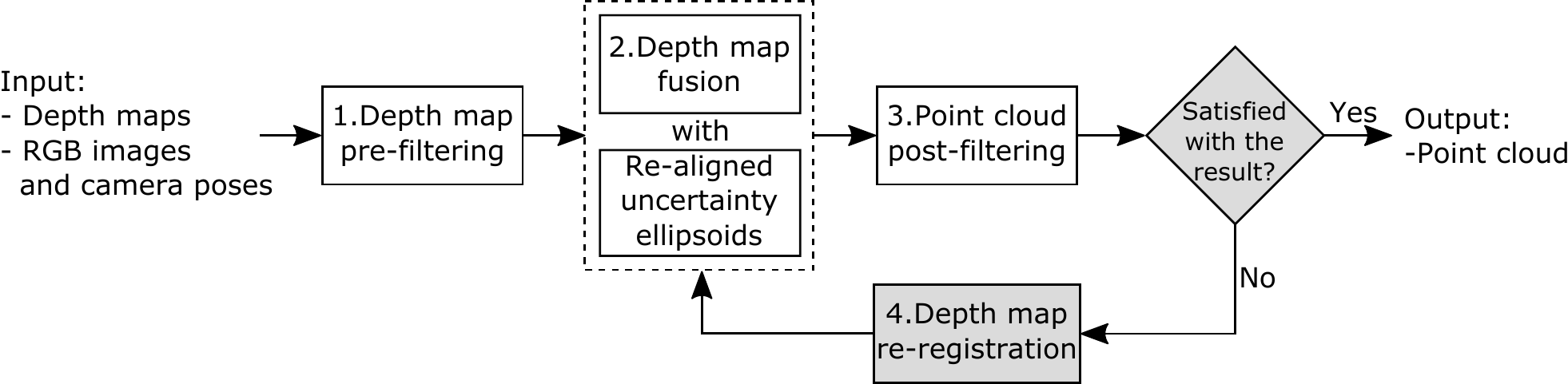}
\caption{An overview of the proposed fusion pipeline. In this paper, we propose the re-registration extension (parts with a gray background) to the fusion algorithm.}
\label{pipeline_overview}
\end{figure}

An overview of the proposed method is presented in Figure \ref{pipeline_overview}. As shown in the figure, the method takes a set of depth maps and RGB images with initial camera positions as input and outputs a point cloud. The method improves the approach in \cite{Ylimaki17} with the re-registration extension marked with darker boxes in the figure. Similarly to \cite{Kyostila13} and \cite{Ylimaki17}, the proposed method can be used as a pipeline to process one depth map at a time so that the new depth maps are re-registered with the overlapping areas in the current point cloud, similarly as in KinectFusion \cite{Newcombe11}. Thus, the only thing that limits the scale or size of the reconstruction is the available memory for storing the point cloud.

The pipeline has totally four steps: 1) depth map pre-filtering, 2) actual depth map fusion, 3) post-filtering of the final point cloud and 4) re-registration of the depth maps into the fused point cloud. The first three steps are briefly described in Section \ref{fusion}. The fourth step is described in more detail in Section \ref{registration}.

\subsection{Fusion pipeline} \label{fusion}

The proposed reconstruction pipeline consists of the depth map fusion and the re-registration. This section shortly introduces the fusion part of the pipeline. 

As an overview, the depth map fusion provides three different ways to measure the uncertainties of the depth measurements and tries to replace, remove or refine the most uncertain ones with better measurements from the other overlapping depth maps.

The first phase in the pipeline is the depth map pre-filtering step, which tries to remove such measurements from the depth maps which seem to be outliers or too inaccurate. Inaccuracy is typically caused by the lens distortion (especially in the corners of the image) or by a phenomenon called multi-path interference (MPI) \cite{Naik15}. MPI errors typically occur in the depth measurements which are acquired with a time-of-flight depth sensor such as Kinect V2 and it causes positive biases to the measurements. MPI happens when the depth sensor receives multiple scattered or reflected signals from the scene for the same pixel. In addition, the depth maps usually have outliers and inaccurate points near depth edges and in dark areas which absorb the majority of the infrared light emitted by the sensor. 

The filtering is based on the observation, that the density of points in a backprojected depth map near outliers and inaccurate or MPI distorted points, is usually smaller than in other, more accurately measured regions. That is, the filtering compares the distances between every backprojected depth measurement and its nearest neighbors to a corresponding reference distance, and removes the measurement if the distance is longer than a certain threshold. In this work, the measurement $m$ is removed if
\begin{equation}
d_m > \gamma d_r(z_m),
\end{equation}
where $d_m$ is the distance from the point $m$ to its 4th nearest neighbor in the measured backprojected depth map, $d_r(z_m)$ is the corresponding reference distance at depth $z_m$ and $\gamma$ is constant variable ($\gamma=1.83$ in our experiments). The reference distance is a function of the depth telling the average distance from a point to its 4th nearest neighbor in a planar point cloud obtained by backprojection of a depth map which has the same depth value in every pixel (e.g. $z_m$ for point $m$).



The second phase in the depth map fusion does the actual fusion based on the depth dependent uncertainty of the measurements. The uncertainty is based on the measurement variances of the used depth sensor. That is, each measurement has a covariance matrix
\begin{equation*} \label{covariance_mat}
\textbf{C} = \begin{bmatrix} 
				{\lambda}_1(\frac{\beta_xz}{\sqrt{12}})^2 & 0 & 0 \\
				0 & {\lambda}_1(\frac{\beta_yz}{\sqrt{12}})^2 & 0 \\
				0 & 0 & {\lambda}_2(\alpha_2z^2+\alpha_1z+\alpha_0)^2
             \end{bmatrix},
\end{equation*}
which represent the location uncertainty of a point in $x$, $y$ and $z$ directions in the 3-D space as depth dependent variances. The parameters $\lambda_1$, $\lambda_2$, are used to scale the variances, $\beta_x$, $\beta_y$, define the width and height of a back projected pixel at one meter away from the sensor and $\alpha_2$, $\alpha_1$ and $\alpha_0$ describe a quadratic depth variance function \cite{Herrera12}. The parameters were calibrated as described in \cite{Ylimaki17} and \cite{Kyostila13}. The uncertainty $\textbf{C}$ defines an ellipsoid in the 3-D space which is aligned so that the z-axis is parallel to the line-of-sight, i.e. the line from the camera center to the point in the space.

\begin{figure}[t!]
\centering
\includegraphics[width=0.7\columnwidth]{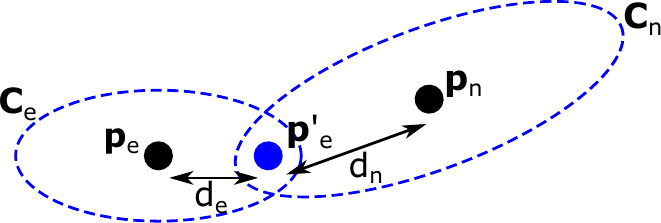}
\caption{Fusion of two nearby depth measurements. $\textbf{C}_e$ and $\textbf{p}_e$ are the covariance and location of the existing measurement. $\textbf{C}_n$ and $\textbf{p}_n$ are the covariance and location of the new measurement. These two measurements are merged together to a point $p'_e$ if the merged point is inside both covariance regions. The point with lower uncertainty (here $\textbf{p}_e$) gets a bigger weight in the refinement ($d_e < d_n$).}
\label{point_refinement}
\end{figure}

Now, when fusing the depth maps, the uncertainties of overlapping points from different depth maps determine whether the points are merged together or not. If the points will be merged together, the new measurement is used to refine the location of the existing measurement. As shown in Figure \ref{point_refinement}, the measurement which seems to be more certain gets a bigger weight, i.e. the refined point $\textbf{p}'_e$ is nearer to the point with lower uncertainty ($d_e < d_n$). If there is no existing measurement near enough, the new measurement is simply added to the cloud of points.

Finally, the third phase filters out the points from the final point cloud appearing in locations which are unlikely based on the statistics collected during the fusion. The statistics consists of the number of merges and the visibility violations of every point. A visibility violation is a vote for a point to be an outlier. The bigger the number of merges the more certainly the point belongs to the reconstruction. Therefore, the points whose visibility violation count is bigger than the count of merges are removed from the cloud at the end. That is, if two nearby points in the space, which were not merged together during the fusion, project into the same pixel in a depth map and both should be visible (i.e. their normals point toward the same half space where the camera is located) then the other point violates the visibility of the other, and thus, is more likely an outlier.

In this work, two points violate the visibility of each other if
\begin{equation}
\arccos(\textbf{n}_e \bullet \textbf{v}_e) < \frac{\pi}{2}  \text{ , } \arccos(\textbf{n}_n \bullet \textbf{v}_n) < \frac{\pi}{2} \text{ and }
\end{equation}
\begin{equation}
|s_e - s_n| < 0.1s_n
\end{equation}
where $\bullet$ is the dot product, $\textbf{n}_e$ and $\textbf{n}_n$ are the normals of the existing and new measurement, respectively, $\textbf{v}_e$ and $\textbf{v}_n$ are normalized vectors from the two points towards the camera and $s_e$ and $s_n$ are the distances between the camera center and the existing and new point, respectively.





\subsection{Pose refinement via re-registration of the depth maps} \label{registration}

In a generic reconstruction pipeline, the initial registration of depth maps or RGB images is based on solving the structure from motion (SfM) problem followed by a bundle adjustment. SfM is based on the tracking of movements of relatively sparse sets of feature points between images, and thus, its accuracy depends on the image content. Therefore, in complicated environments having repetitive textures, the initial registration usually has room for improvement.

As shown in Figure \ref{pipeline_overview}, our method incorporates a re-registration step into the reconstruction pipeline. The registration aligns the original depth maps with the last fused point cloud. It is based on the iterative closest point (ICP) algorithm, which iteratively refines the given extrinsic parameters of a depth camera to minimize the distance between the corresponding backprojected depth map and the fused point cloud. That is, let denote $\textbf{R}_i$ and $\textbf{t}_i$ the initial rotation and translation of the \textit{ith} camera in relation to the global coordinate frame and $\textbf{D}_{xi}' = \textbf{R}_i^{T}\textbf{D}_{xi} - \textbf{R}_i^T\textbf{t}_i$, the backprojected pixel $x$ of depth map $i$ in the global coordinate frame, where $\textbf{D}_{xi}$ are the coordinates of the corresponding pixel in the coordinate frame of the \textit{ith} camera. Now, the method iteratively tries to find a rotation $\hat{\textbf{R}}_i$ and a translation $\hat{\textbf{t}}_i$ for each camera \textit{i} so that the error between the points $\hat{\textbf{R}}_i\textbf{D}'_{xi} + \hat{\textbf{t}}_i, \forall x$ and the points in the full fused point cloud is minimized \cite{Chen92}. After $m$ iterations ($m$ = 10 in our experiments), the camera poses are updated with $\textbf{R}_i \leftarrow \hat{\textbf{R}}^{T}_i\textbf{R}_i$ and $\textbf{t}_i \leftarrow \hat{\textbf{R}}^{T}_i\textbf{R}_i - \hat{\textbf{R}}^{T}_i\hat{\textbf{t}}_i$ and the fusion step is repeated as illustrated in Figure \ref{pipeline_overview}. The registration and fusion steps can be repeated until the result is satisfying or does not get significantly better between iterations. In our experiments, the pipeline was iterated six times. The re-registration was implemented with C++ using the Point Cloud Library\footnote{http://pointclouds.org/} (PCL).


\section{Experimental results} \label{results}

\begin{figure}[t!]
\centering
\includegraphics[width=0.32\columnwidth]{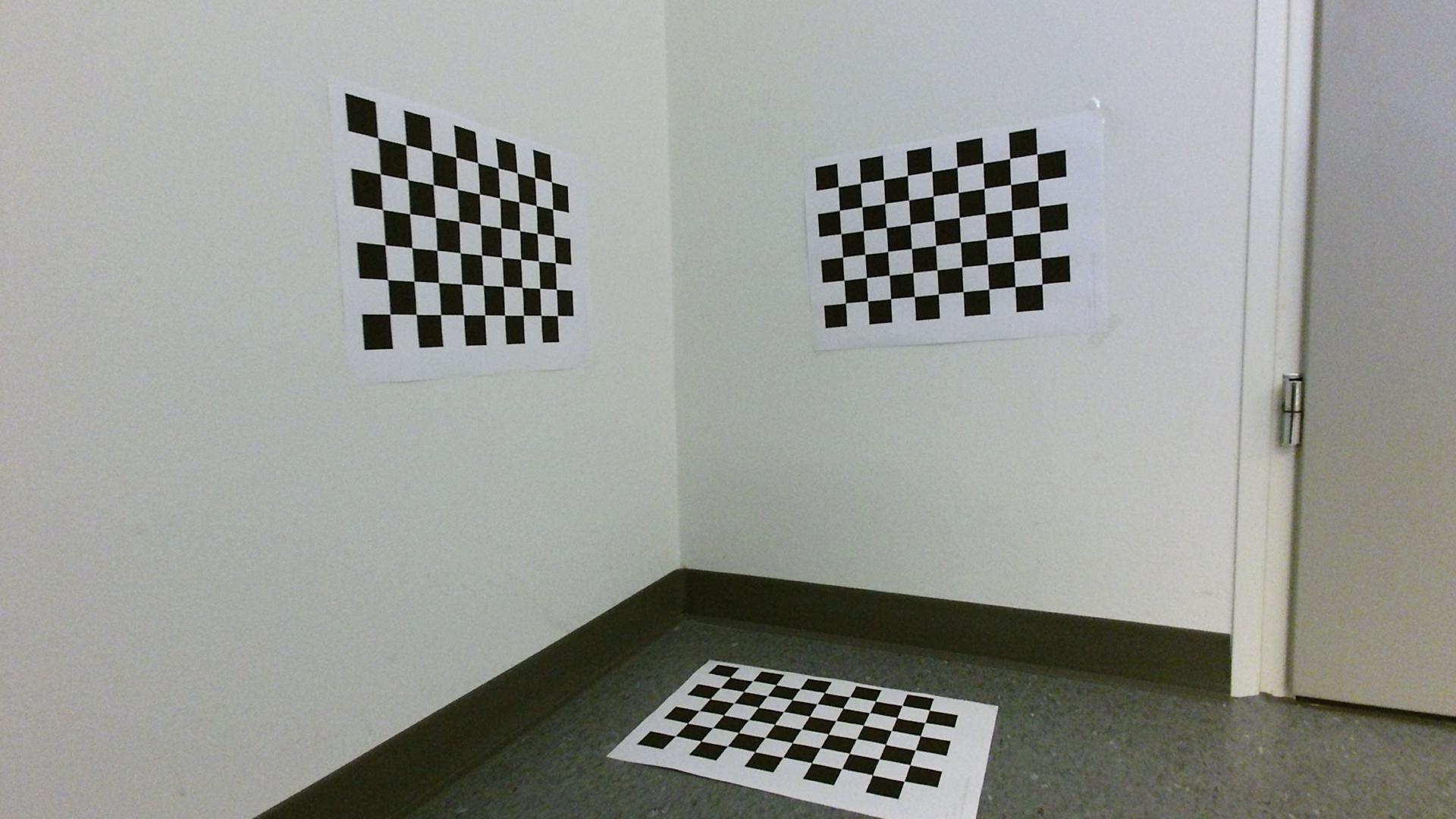}
\includegraphics[width=0.32\columnwidth]{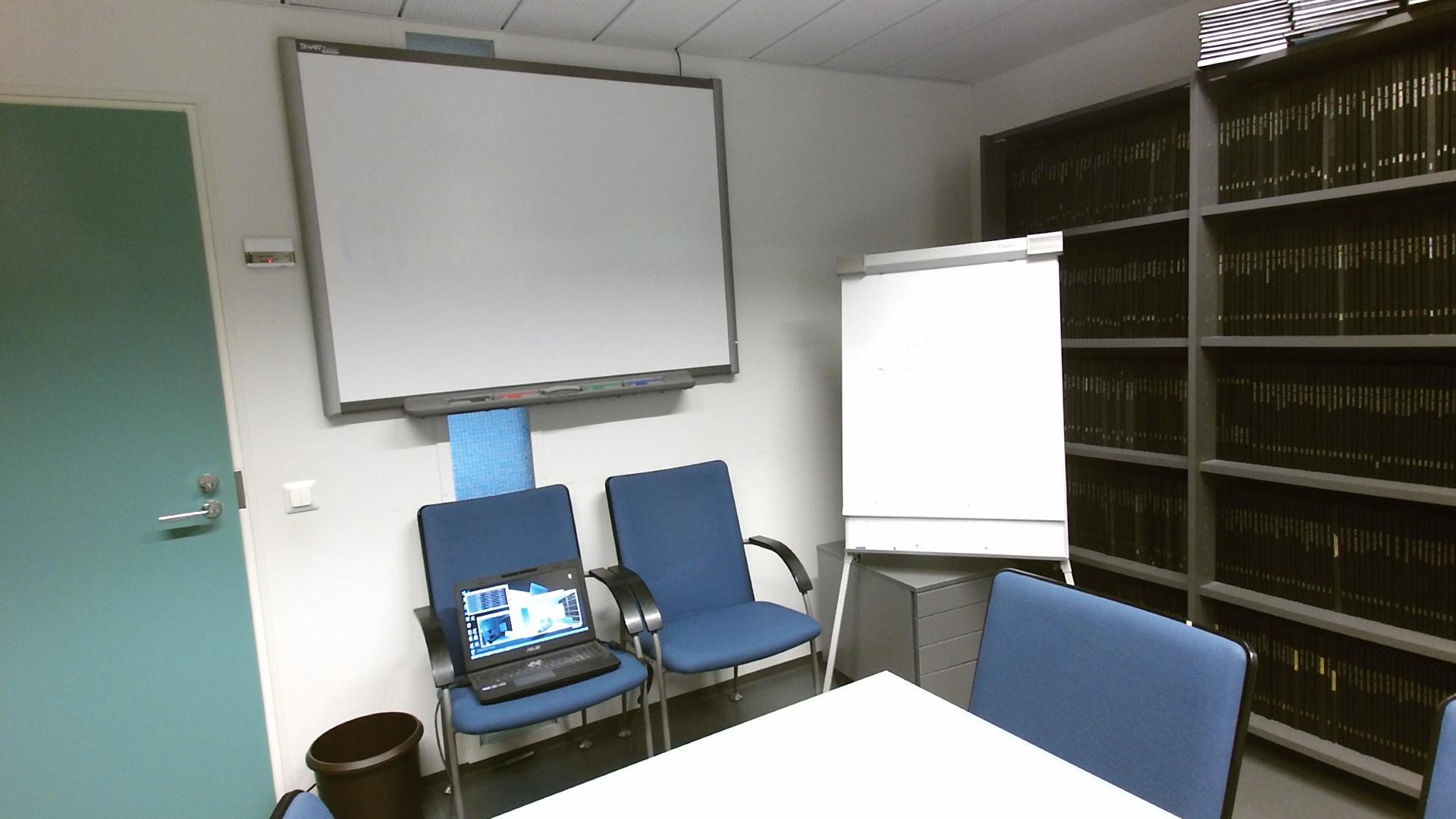}
\includegraphics[width=0.32\columnwidth]{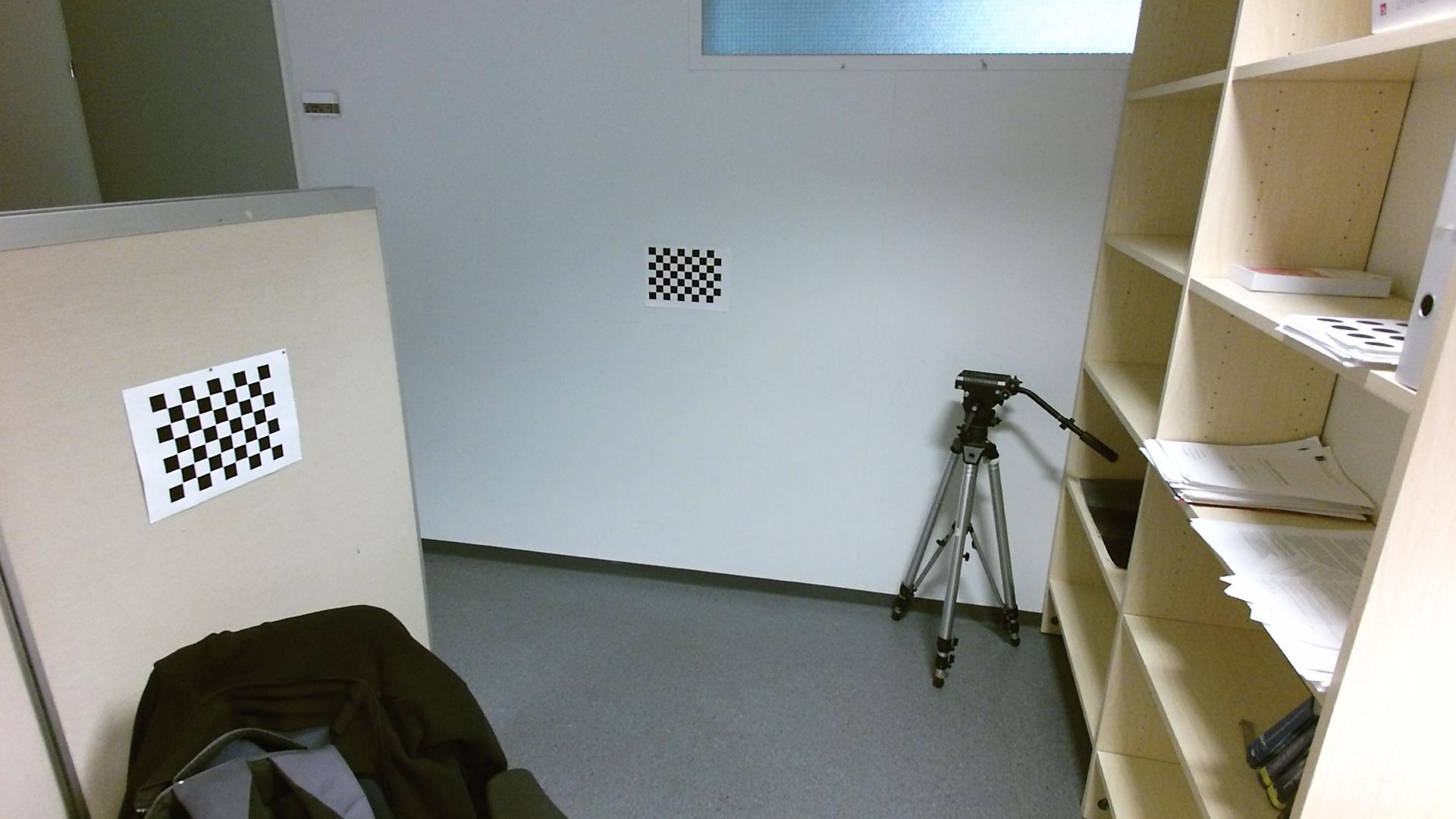}
\caption{A sample image from each dataset used in the experiments. From left to right: CCorner, Office1 and Office2.}
\label{sample_imgs}
\end{figure}

The experiments were made using three datasets (i.e. CCorner, Office1 and Office2), captured with Kinect V2. Figure \ref{sample_imgs} illustrates a sample image from each dataset. The first dataset is a simple concave corner whereas the other two are rather complicated office environments. The results made with the proposed method were compared with the results made with the methods in \cite{Kyostila13} and \cite{Ylimaki17}. The experiments include both visual and quantitative evaluations of the obtained results.

\subsection{Visual evaluation of the results}

\begin{figure*}
\centering
\includegraphics[width=0.31\textwidth]{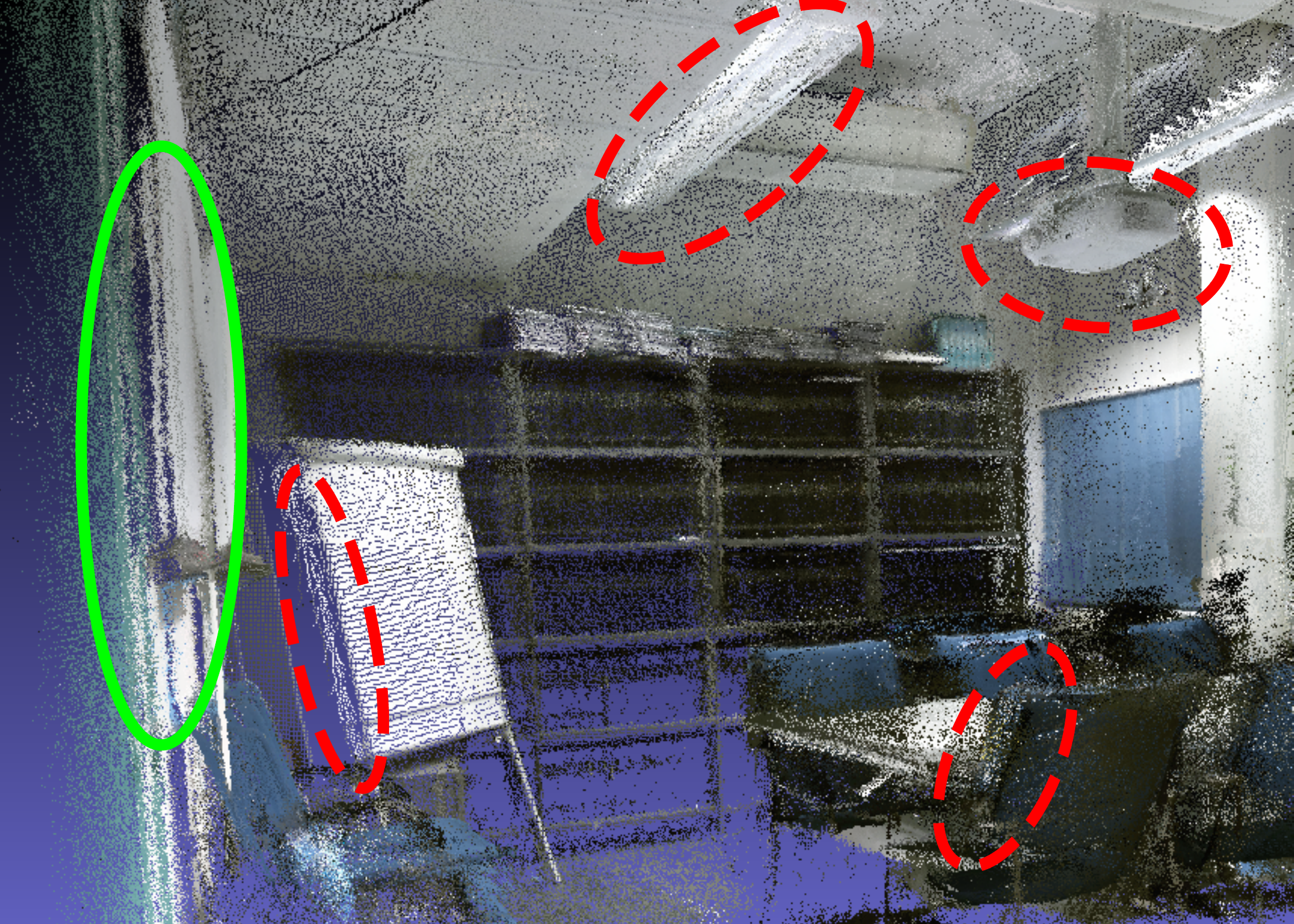}
\includegraphics[width=0.31\textwidth]{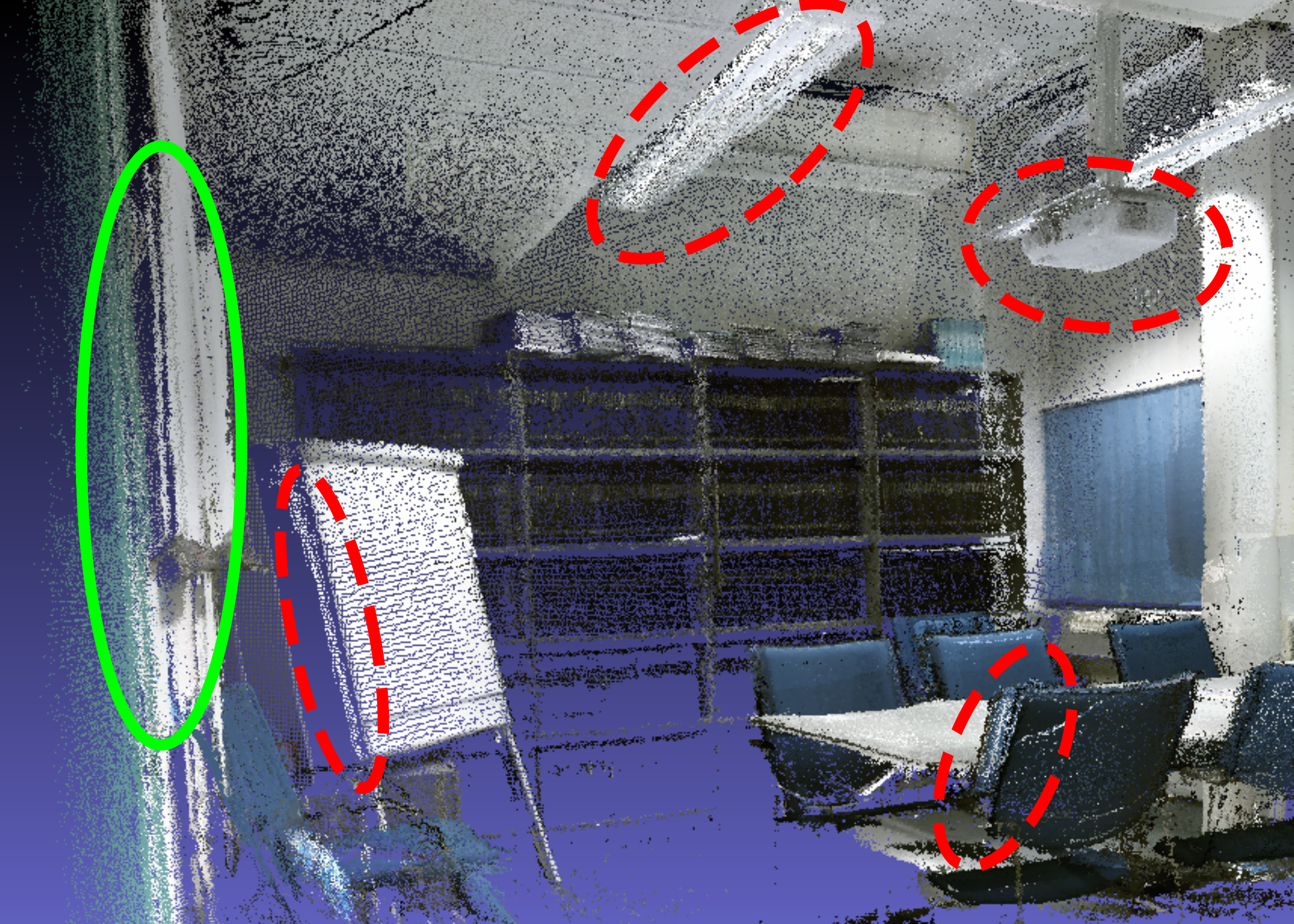}
\includegraphics[width=0.31\textwidth]{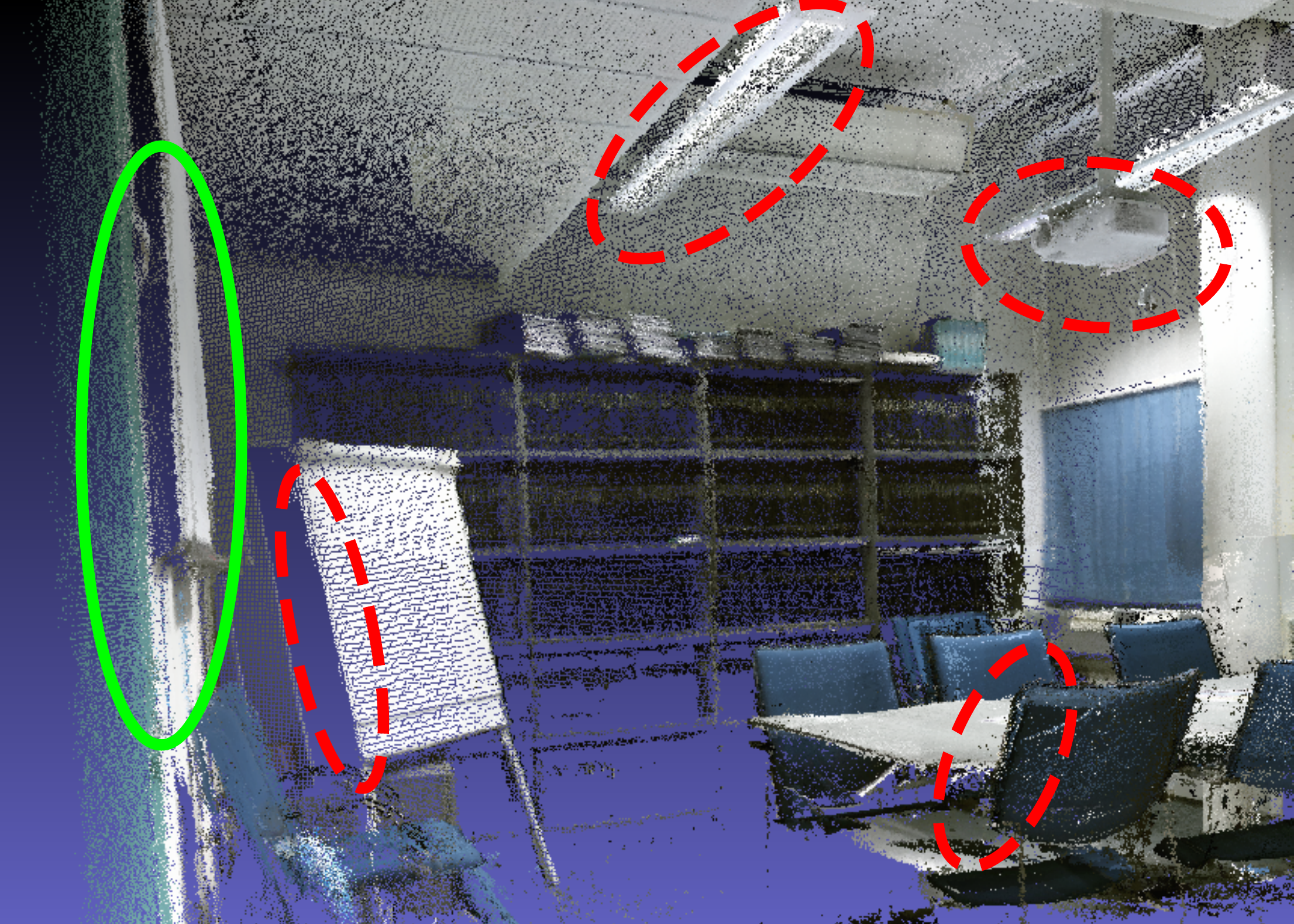} \\
\vspace{2mm}
\includegraphics[width=0.31\textwidth]{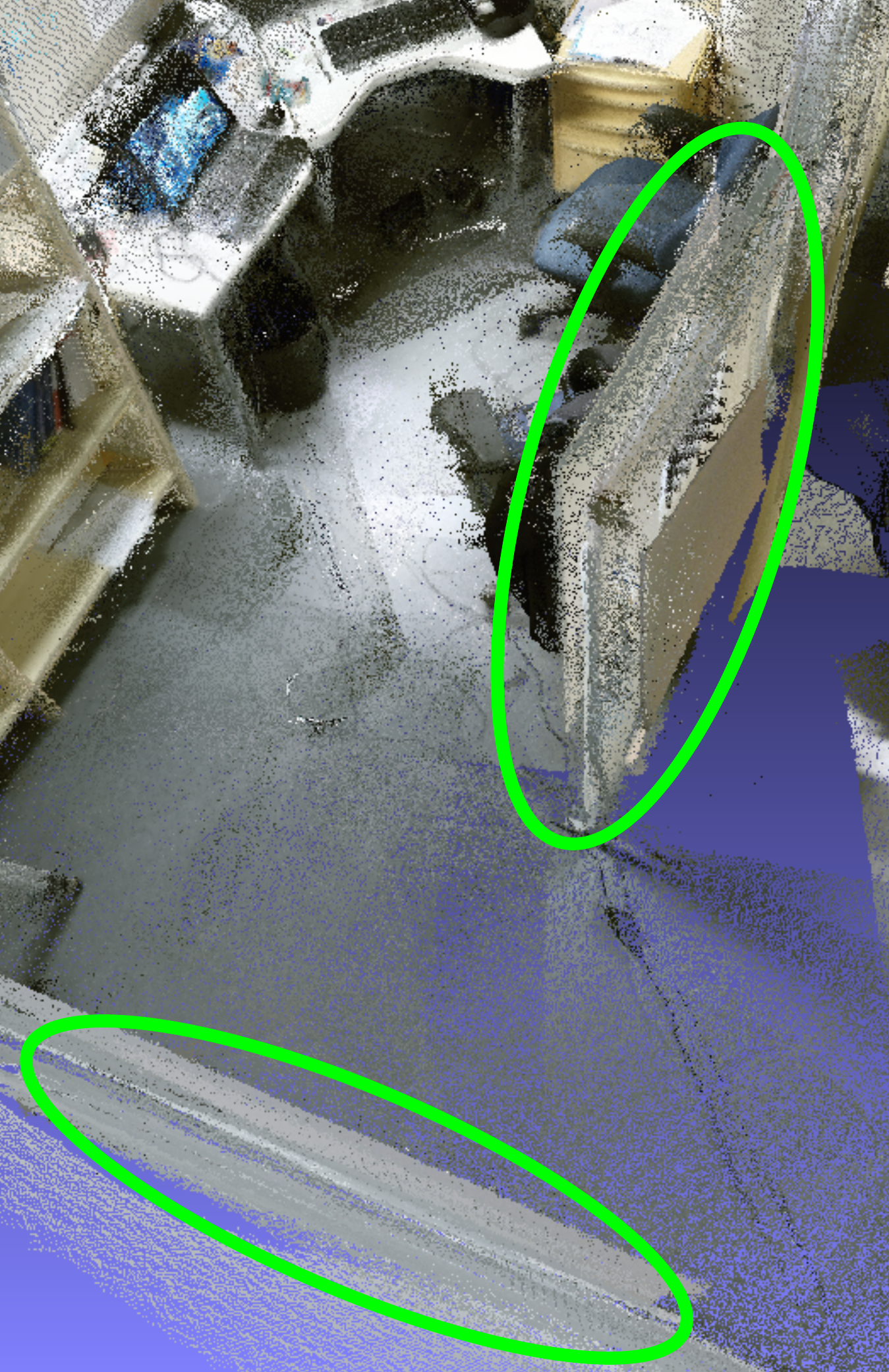}
\includegraphics[width=0.31\textwidth]{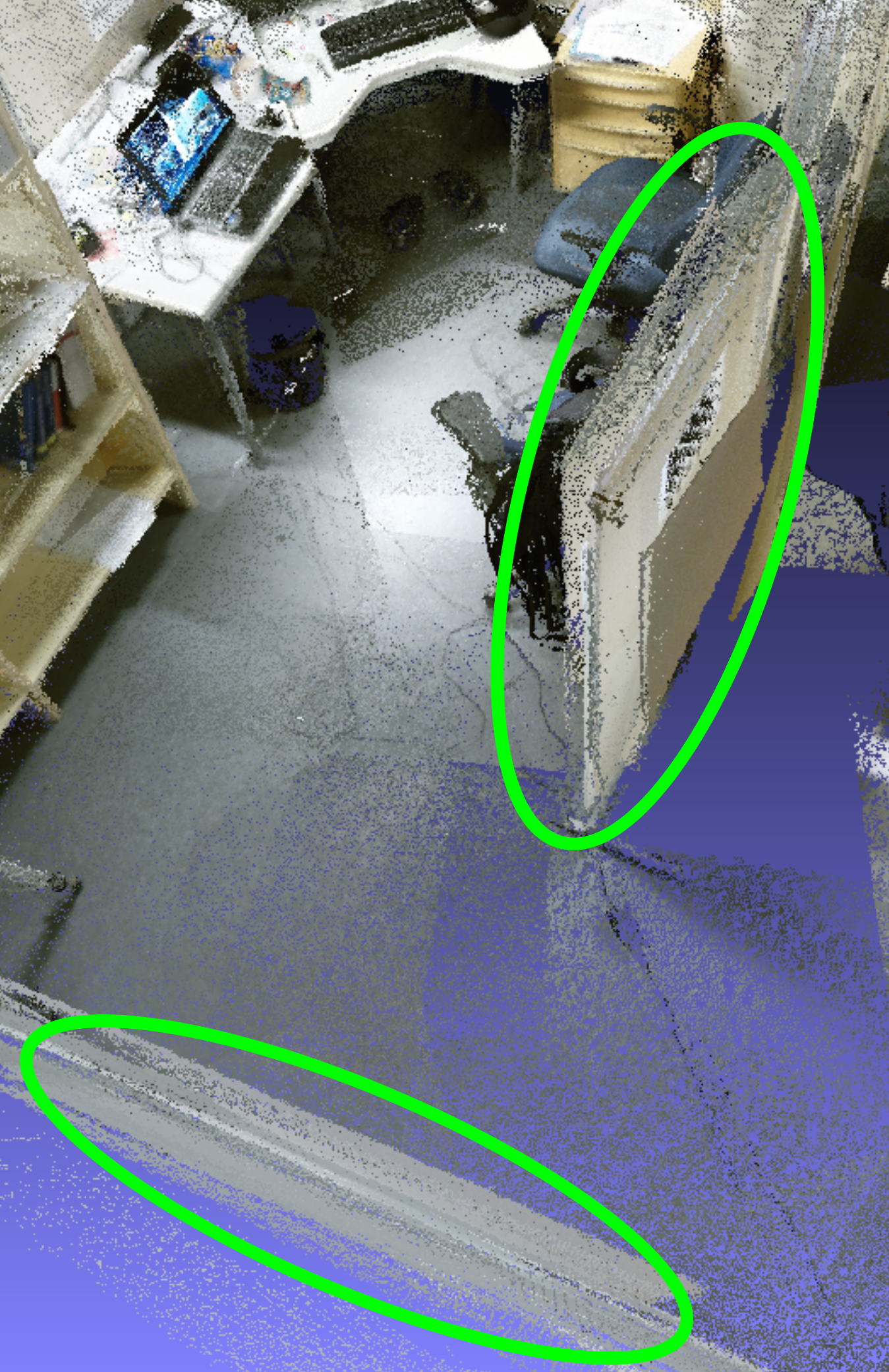}
\includegraphics[width=0.31\textwidth]{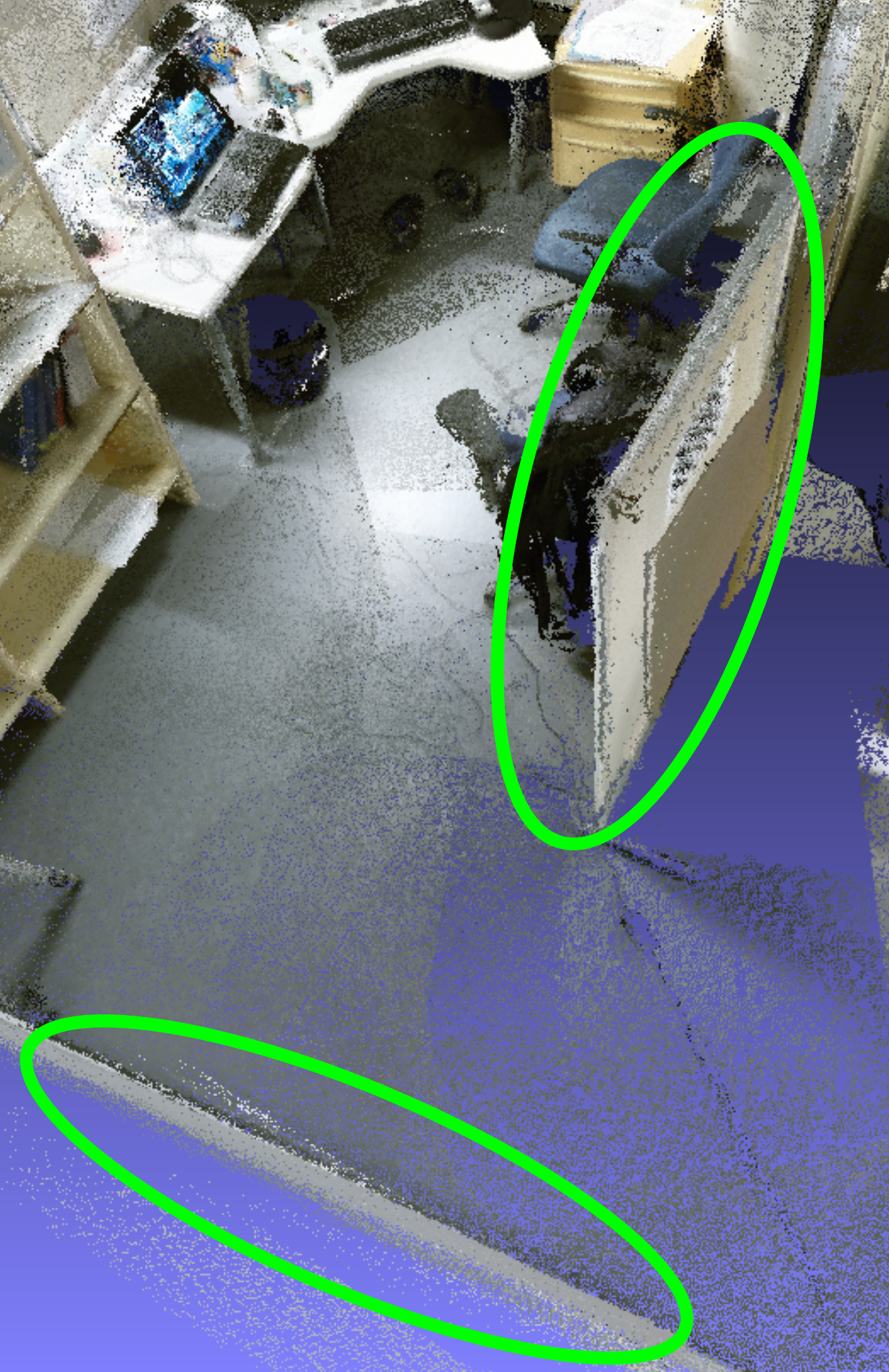} 
\caption{Comparison between Office1 (top) and Office2 (bottom) results made with the methods in \cite{Kyostila13} (left) and \cite{Ylimaki17} (middle) and with the method proposed in this paper (right). The proposed method significantly reduces the amount of registration errors which makes the surfaces less ambiguous (green ellipses) and the object boundaries more accurate (red dashed ellipses).}
\label{kyostila_vs_scia_vs_ours}
\end{figure*}

In the first experiment, we compared the results acquired from Office1 and Office2 datasets. Figure \ref{kyostila_vs_scia_vs_ours} presents the point clouds made with \cite{Kyostila13}, \cite{Ylimaki17} and the proposed method. The figure clearly shows that, compared with \cite{Kyostila13}, the method in \cite{Ylimaki17} is able to reduce the amount of outliers but it cannot correct the registration errors. The registration errors cause ambiguous surfaces (green ellipses) and inaccurate object boundaries (red dashed ellipses), for examples, which do not appear in the results made with the proposed method, as shown on the right in Figure \ref{kyostila_vs_scia_vs_ours}.


\subsection{Quantitative evaluations}

In the second experiment, the results obtained with the methods in \cite{Kyostila13}, \cite{Ylimaki17} and the proposed one were compared quantitatively in three ways. First, Table \ref{stats_table} presents the sizes of the used datasets and the sizes of the final point clouds. As shown, the proposed method decreases the number of points in the final point cloud.


\begin{table}[b!]
\centering
\vspace{-2mm}
\caption{An overview of the statistics of the datasets.}
\label{stats_table}
\begin{tabular}{|l|c|c|c|c|c|}
\hline
\textbf{\scriptsize{Dataset}} & \textbf{\begin{tabular}[c]{@{}c@{}}\scriptsize{View}\\ \scriptsize{count}\end{tabular}} & \textbf{\begin{tabular}[c]{@{}c@{}}\scriptsize{Original}\\ \scriptsize{point count}\end{tabular}} & \textbf{\scriptsize{Method}} & \textbf{\begin{tabular}[c]{@{}c@{}}\scriptsize{Final}\\ \scriptsize{point count}\end{tabular}} & \textbf{\begin{tabular}[c]{@{}c@{}}\scriptsize{Ratio of}\\ \scriptsize{reduction}\end{tabular}} \\
\hline
\multirow{3}{*}{\textbf{\scriptsize{CCorner}}} & \multirow{3}{*}{\scriptsize{59}} & \multirow{3}{*}{\scriptsize{9 307 296}} & \scriptsize{\cite{Kyostila13}} & \scriptsize{1 299 555} & \scriptsize{86.0\%} \\
\cline{4-6}
& & & \scriptsize{\cite{Ylimaki17}} & \scriptsize{939 730} & \scriptsize{89.9\%} \\
\cline{4-6}
& & & \scriptsize{Ours} & \scriptsize{881 994} & \scriptsize{90.5\%} \\
\hline
\multirow{3}{*}{\textbf{\scriptsize{Office1}}} & \multirow{3}{*}{\scriptsize{98}}  & \multirow{3}{*}{\scriptsize{16 690 662}} & \scriptsize{\cite{Kyostila13}} & \scriptsize{5 930 663} & \scriptsize{64.5\%} \\
\cline{4-6}
& & & \scriptsize{\cite{Ylimaki17}} & \scriptsize{4 352 962} & \scriptsize{73.9\%} \\
\cline{4-6}
& & & \scriptsize{Ours} & \scriptsize{4 252 937} & \scriptsize{74.5\%} \\
\hline
\multirow{3}{*}{\textbf{\scriptsize{Office2}}} & \multirow{3}{*}{\scriptsize{114}} & \multirow{3}{*}{\scriptsize{20 400 588}} & \scriptsize{\cite{Kyostila13}} & \scriptsize{6 777 222} & \scriptsize{66.7\%} \\
\cline{4-6}
& & &\scriptsize{\cite{Ylimaki17}} & \scriptsize{5 221 117} & \scriptsize{74.4\%} \\
\cline{4-6}
& & & \scriptsize{Ours} & \scriptsize{4 956 266} & \scriptsize{75.7\%} \\
\hline
\end{tabular}
\end{table}

\begin{figure}[t!]
\centering
\includegraphics[width=0.8\columnwidth]{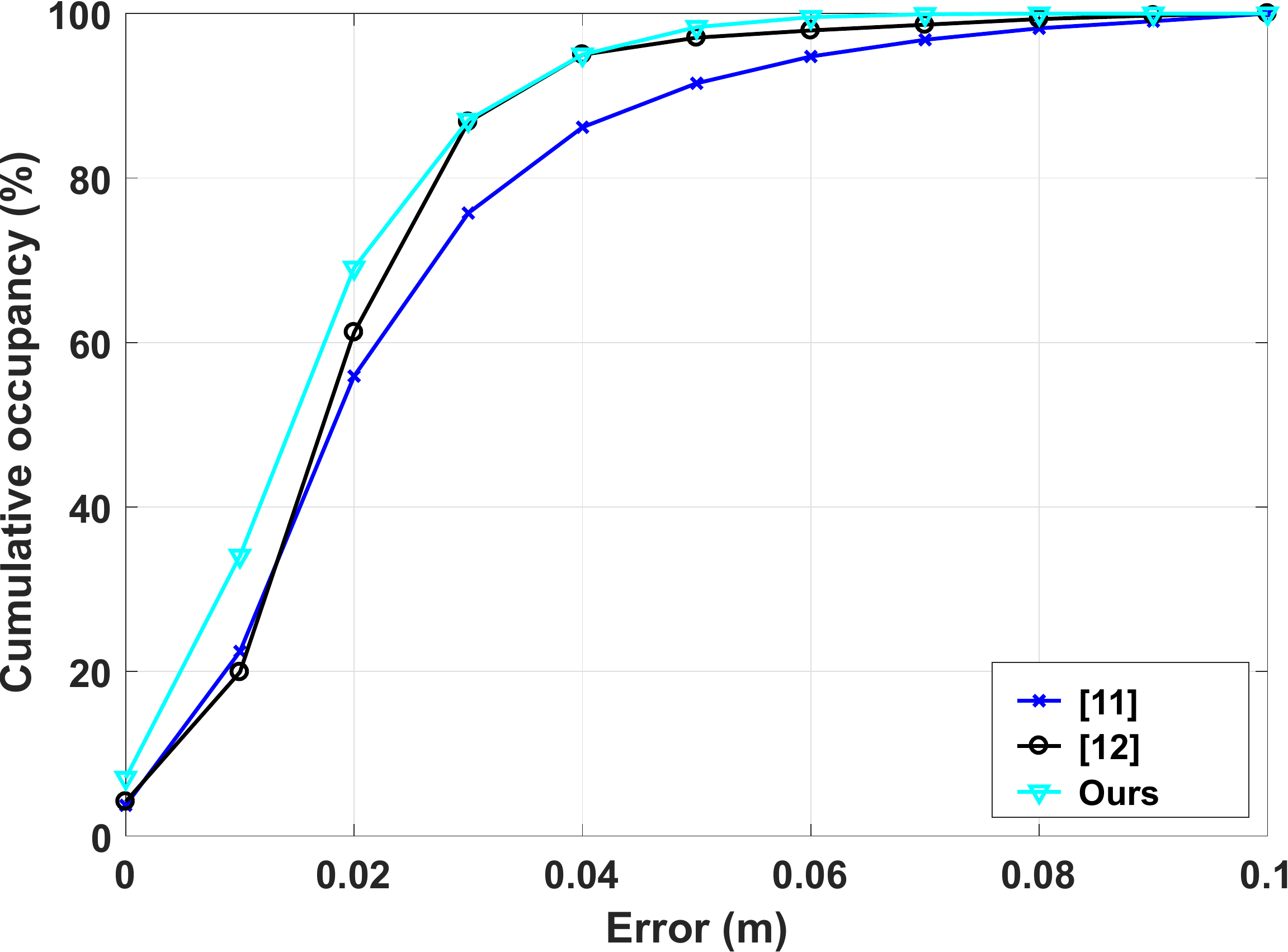}
\vspace{-1mm}
\caption{Evaluation of the leftover errors in the CCorner reconstructions made with \cite{Kyostila13}, \cite{Ylimaki17} and the proposed method.}
\vspace{-5mm}
\label{ccorner_errors}
\end{figure}

Then, the reconstructions of CCorner were evaluated in two ways by comparing them against a ground truth which consists of three orthogonal planes. In the first evaluation, the errors between the point clouds and the ground truth were obtained by calculating distances from every point to the nearest ground truth plane (i.e. floor, right wall or left wall). The errors are presented in Figure \ref{ccorner_errors} as a cumulative curve where the value on the y-axis is a percentage of points whose error is smaller or equal to the corresponding value on the x-axis.

As shown in the figure, the proposed method enhances the accuracy of the model. That is, the proportions of points whose error is below or equal to 0, 0.1 or 0.2, are clearly bigger than the corresponding values of the other two methods.

Then, CCorner point clouds were evaluated using the voxel based evaluation metric proposed in \cite{Ylimaki15b}. The evaluation values are presented in Table \ref{voxel_eval}. The evaluation values indicate the completeness and compactness of the reconstructions. The completeness is defined with Jaccard index which indicates the proportion of the ground truth which is covered by the reconstruction within a certain threshold. Jaccard index is calculated by comparing the voxel representation of the reconstruction, and thus, the threshold, mentioned above, is the size of the voxel (i.e. length of an edge of a voxel). The compactness, instead, is a compression ratio calculated as a ratio of the number of the points in the ground truth and the reconstruction.

As the table shows, the point cloud achieved with the proposed method is both more compact (bigger compression ratio) and complete (bigger Jaccard index regardless of the size of a voxel) than the reconstructions made with \cite{Kyostila13} and \cite{Ylimaki17}. 


\begin{table}[t!]
\centering
\caption{Evaluation of completeness (Jaccard index) and compactness (compression ratio) of CCorner reconstructions \cite{Ylimaki15b}. See the text for the details.} 
\vspace{-1mm}
\label{voxel_eval}
\begin{tabular}{|l|l|c|c|c|}
\hline
\multicolumn{2}{|l|}{\multirow{2}{*}{}} & \multicolumn{3}{c|}{\textbf{Method}} \\ 
\cline{3-5} 
\multicolumn{2}{|l|}{} & {\cite{Kyostila13}} & {\cite{Ylimaki17}} & Ours \\ 
\hline
\multicolumn{2}{|l|}{\textbf{Compression ratio}} & 0.443 & 0.612 & {\ul 0.652} \\ 
\hline
\multirow{4}{*}{\textbf{\begin{tabular}[c]{@{}l@{}}Jaccard index\\ with voxel size\end{tabular}}} 
& \textbf{5mm} & 0.027 & 0.026 & {\ul 0.045} \\ 
\cline{2-4} 
& \textbf{20mm} & 0.162 & 0.174 & {\ul 0.190} \\ 
\cline{2-4} 
& \textbf{45mm} & 0.309 & 0.350 & {\ul 0.355} \\ 
\cline{2-4} 
& \textbf{85mm} & 0.388 & 0.440 & {\ul 0.456} \\ 
\hline
\end{tabular}
\vspace{-3mm}
\end{table}

\subsection{Discussion}

The results showed that the proposed method produced significantly better results than its predecessors. The visual comparisons indicated that the point clouds made with the proposed method are cleaner and less ambiguous than those made with \cite{Kyostila13} and \cite{Ylimaki17}. Although the method in \cite{Ylimaki17} is able to remove the majority of the incorrect depth measurements, which appear in \cite{Kyostila13}, it cannot properly handle points which have been registered incorrectly. Therefore, the re-registration extension is an important improvement to the reconstruction pipeline. 

The re-registration of the proposed method improves the registration of a depth map if at least part of its points overlap with points in other depth maps so that they have been fused together during the fusion phase. In the fusion phase, as described earlier, a new measurement is used to refine a nearby existing point if there is any or it is added to the cloud otherwise. If most of the measurements of a depth map are only added to the cloud, the re-registration is not able to properly refine the alignment of the depth map because the error between the added points and the original map is zero. However, such situations could be avoided by capturing the scene more carefully, and thus, ensuring that the depth maps have enough redundancy and the initial camera poses can be calculated relatively well.

The quantitative evaluations showed that the reconstructions obtained with the proposed method contained less points than the reconstructions made with \cite{Kyostila13} and \cite{Ylimaki17}. Nevertheless, the completeness of the models even increased slightly. That is, the proposed method is able to decrease the redundancy without decreasing the completeness.


\section{Conclusion} \label{conclusion}

In this paper, we proposed a method for depth map fusion. The proposed method merges a sequence of depth maps into a single non-redundant point cloud. The fusion pipeline consists of the actual depth map fusion and a re-registration phase which are iterated until the result is satisfying or does not change significantly. The fusion phase gets the depth maps and corresponding camera poses as input and produces a non-redundant point cloud. The re-registration phase instead, tries to refine the original poses of the cameras by the registration of the original backprojected depth maps into the fused point cloud. Then, the depth maps and refined camera poses are fed again to the fusion phase. The experiments showed that the proposed method is able to produce more accurate and unambiguous reconstructions than its predecessors.

\bibliographystyle{IEEEtran}
\bibliography{references}
%



\end{document}